\documentclass[runningheads,a4paper]{llncs}

\usepackage{amssymb}
\usepackage{graphicx}

\usepackage{url}
\urldef{\mailsa}\path|{alfred.hofmann, ursula.barth, ingrid.haas, frank.holzwarth,|
\urldef{\mailsb}\path|anna.kramer, leonie.kunz, christine.reiss, nicole.sator,|
\urldef{\mailsc}\path|erika.siebert-cole, peter.strasser, lncs}@springer.com|    
\newcommand{\keywords}[1]{\par\addvspace\baselineskip
\noindent\keywordname\enspace\ignorespaces#1}

\begin{document}

\mainmatter  % start of an individual contribution

\title{Regression Forest-based Atlas Localization and Direction Specific Atlas Generation for \\Pancreas Segmentation}

\titlerunning{Atlas Localization and Atlas Generation for Pancreas Segmentation}

\author{Masahiro Oda \inst{1}
% \thanks{The authors thank our colleagues for suggestions and advices. Parts of this research were supported by the Grant-in-Aid for Scientific Research from the Ministry of Education (MEXT), Japan Society for the Promotion of Science (JSPS), the Fund for Cancer Research and Development from the National Cancer Center, and the Kayamori Foundation of Informational Science Advancement.}%
\and
Natsuki Shimizu \inst{1}
\and
Ken'ichi Karasawa \inst{1}
\and
Yukitaka Nimura \inst{2}
\and
Takayuki Kitasaka \inst{3}
\and
Kazunari Misawa \inst{4}
\and
Michitaka Fujiwara \inst{5}
\and
\\
Daniel Rueckert \inst{6}
\and
Kensaku Mori \inst {2,1}}

\authorrunning{Masahiro Oda et al.}

% \author{***** ***** \inst{1}
% \and
% ***** ***** \inst{1}
% \and
% ***** ***** \inst{1}
% \and
% ***** ***** \inst{2}
% \and
% ***** ***** \inst{3}
% \and
% ***** ***** \inst{4}
% \and
% ***** ***** \inst{5}
% \and
% ***** ***** \inst{6}
% \and
% ***** ***** \inst {7,1}}

% \authorrunning{***** ***** et al.}

\institute{Graduate School of Information Science, Nagoya University,\\
%Furo-cho, Chikusa-ku, Nagoya, Aichi, 464-8601, Japan\\
\email{moda@mori.m.is.nagoya-u.ac.jp}\\
\and
Strategy Office, Information and Communications, Nagoya University,\\
%Furo-cho, Chikusa-ku, Nagoya, Aichi, 464-8601, Japan
\and
School of Information Science, Aichi Institute of Technology,\\
%1247 Yachigusa, Yagusa-cho, Toyota, Aichi, 470-0392, Japan
\and
Aichi Cancer Center,\\
\and
Nagoya University Graduate School of Medicine,\\
\and
Department of Computing, Imperial College London
%\url{http://www.mori.m.is.nagoya-u.ac.jp/~moda/index-e.html}
}

% \institute{*****\\
% \and
% ***** \\
% \and
% ***** \\
% \and
% ***** \\
% \and 
% ***** \\
% \and 
% ***** \\
% \and 
% *****
% }


\toctitle{Lecture Notes in Computer Science}
\tocauthor{Authors' Instructions}
\maketitle

%The abstract should summarize the contents of the paper and should contain at least 70 and at most 150 words. It should be written using the \emph{abstract} environment.
\begin{abstract}
This paper proposes a fully automated atlas-based pancreas segmentation method from CT volumes utilizing atlas localization by regression forest and atlas generation using blood vessel information.
Previous probabilistic atlas-based pancreas segmentation methods cannot deal with spatial variations that are commonly found in the pancreas well.
Also, shape variations are not represented by an averaged atlas.
We propose a fully automated pancreas segmentation method that deals with two types of variations mentioned above.
The position and size of the pancreas is estimated using a regression forest technique.
After localization, a patient-specific probabilistic atlas is generated based on a new image similarity that reflects the blood vessel position and direction information around the pancreas.
We segment it using the EM algorithm with the atlas as prior followed by the graph-cut.
In evaluation results using 147 CT volumes, the Jaccard index and the Dice overlap of the proposed method were 62.1\% and 75.1\%, respectively.
Although we automated all of the segmentation processes, segmentation results were superior to the other state-of-the-art methods in the Dice overlap.
\keywords{segmentation, pancreas, probabilistic atlas, regression forest}
\end{abstract}

\section{Introduction}

Organ segmentation from medical images is an essential process for computer-aided diagnosis and interventions.
%Segmentation results are used for computer aided diagnosis and interventions.
In laparoscopic gastrectomy, pancreas region information is strongly required for enabling safer surgical procedure.
Multi-organ and single-organ segmentation methods for the abdominal region have been proposed\cite{Okada08,Chu13,Wolz13,Karasawa15,Tong15,Saito16}.
High segmentation accuracies were reported for the liver, kidneys, and so on.
Compared to those organs, the segmentation accuracies of the pancreas remain low.

The difficulty of pancreas segmentation results from the large variations of its position and shape.
The pancreas is small, long, and thin.
Its position and shape have a strong relationship to the surrounding organs, including the stomach, the liver, and the intestines.
The pancreas slides and changes its shape and is affected by the surrounding organs.
To segment it, we have to consider two variations: (1) {\it shape} and (2) {\it spatial} (positional variation in the body).
Previously, many probabilistic atlas-based pancreas segmentation methods\cite{Okada08,Chu13,Wolz13,Karasawa15,Tong15} have been developed.
Such atlas-based methods generate a map that represents probability or likelihood of an organ's existence in medical images.
%A single-atlas commonly represents an average shape of the organ obtained from training data.
Patient-specific-atlas\cite{Okada08,Karasawa15,Tong15} and hierarchical-atlas\cite{Wolz13} segmentation methods were developed to deal with the {\it shape variation} of organs.
However, the {\it spatial variation} of the organ is not very well dealt with.
%Atlas-based methods require localization of probabilistic atlases to appropriate locations in medical images.
Previous methods have tried to localize probabilistic atlases using positional information of other organs\cite{Chu13,Wolz13} or manually specified information\cite{Okada08,Karasawa15}.

We propose a fully automated atlas-based pancreas segmentation method.
To deal with the {\it spatial variation} of the pancreas, we used a regression forest technique to estimate its position and size.
%The localization result is used in subsequent atlas-based segmentation.
We construct a patient-specific probabilistic atlas of the pancreas from the CT volumes in a database.
The atlas is constructed for an unknown input CT volume taking the anatomical structures around the pancreas into account.
The pancreas position and shape are related to the surrounding blood vessels.
We use directed structure specific (DSS) volumes, which store blood vessel position and direction information, in the similar volume retrieval of an atlas construction.
In the segmentation process, we segment the pancreas using the expectation maximization (EM) algorithm which uses the atlas as prior and the graph-cut optimization method.

The contributions of this paper are summarized as follows: 
(1) automated pancreas localization based on the relationships between local appearances and pancreas positions; 
(2) patient-specific probabilistic atlas generation of the pancreas taking the blood vessel position and direction information into account; 
and (3) a fully automated framework of pancreas segmentation.

\section{Method}

\subsection{Overview}

Our method segments the pancreas region from an input CT volume.
The pancreas position and size (represented as a bounding box) are estimated by a regression forest technique.
The localization is performed by considering the positional relationships between the local appearances in the CT volume and the pancreas position.
Then we construct a patient-specific probabilistic atlas of the pancreas.
To construct it, we perform CT volume retrieval from a training database, which resembles the unknown input CT volume using directed line structure similarity.
We construct an atlas from the selected volumes and use it for automatic segmentation.

\subsection{Pancreas localization}
\label{ssec:localization}

\subsubsection{Regression forest training}

Our training database includes CT volumes and their corresponding manually segmented pancreas regions.
We defined the axis-aligned minimum bounding box of the pancreas as follows: ${\bf b}=(b_{L}, b_{R}, b_{A}, b_{P}, b_{H}, b_{F})$.
These elements represent the left, right, anterior, posterior, head, and foot side face positions of the bounding box in the CT coordinate system.
We obtain patches of the size of $p \times p \times p$ voxels from the CT volumes on a regular grid.
The patch's center position is defined as ${\bf v}=(v_{x}, v_{y}, v_{z})$.
The patch's offset with respect to ${\bf b}$ is shown as ${\bf d}= {\bf b}-{\bf \hat{v}}$, where ${\bf \hat{v}}=(v_{x}, v_{x}, v_{y}, v_{y}, v_{z}, v_{z})$.

A regression forest is constructed for each face position of the bounding box, which is obtained by constructing six regression forests.
%1 つの木の構築に使用するサンプルの割合はPとする．
During the construction of a regression tree, training patches are pushed to the root of the tree.
At each split node, the thresholding of a feature value of a training patch is performed to decide which child node should be selected for the training patch.
Similar to\cite{Criminisi13}, we use
\begin{equation}
f({\bf v}) = \frac{1}{|F_{1}|} \sum_{{\bf q} \in F_{1}} I({\bf q}) - \frac{1}{|F_{2}|} \sum_{{\bf q} \in F_{2}} I({\bf q})
\label{eq:featurevalue}
\end{equation}
as the feature value of ${\bf v}$.
$I({\bf q})$ is a CT value at voxel ${\bf q}$ in a CT volume.
$F_{1}$ and $F_{2}$ are cuboids arbitrarily arranged in a patch.
$|F_{1}|$ and $|F_{2}|$ denote the number of voxels in the cuboids.
%We select two voxels in a patch ${\bf a}_{1}$ and ${\bf a}_{2}$.
We randomly selected the positions and the sizes of the cuboids in a patch.
Two cuboids may overlap.
%しきい値による左右ノードへの分け方説明

For each split node, we calculate the variance of the offset value
\begin{equation}
\sigma^{2} = \frac{1}{S} \sum^{S}_{n=1}({\bf d}_{n}-{\bf {\bar d}})^{2},
\end{equation}
where $S$ is the number of patches reaching the node, ${\bf d}_{n}$ is the offset in the node, and ${\bf {\bar d}}$ is the average of the offsets in the node.
The variance of a parent node is
\begin{equation}
s^{2} = \sum_{m \in \{L,R\}} \omega_{m} \sigma^{2}_{m},
\end{equation}
where $\omega_{m}=S_{m}/S$.
We iterate the random selection of the feature values $n_{F}$ times and the random selection of the threshold at each node $n_{H}$ times.
A pair of a feature value and a threshold, which gives the minimum variance, is recorded in a split node.
A split node is changed to a leaf node if the number of patches that arrive at the node is fewer than $n_{min}$ or the tree depth exceeds $D$.
In each leaf node, we generate a histogram of patches about their offset.
We estimate the parameters of a normal distribution that fit the histogram.
The EM algorithm is used to estimate the parameters of the normal distribution (mean and variance).

\subsubsection{Bounding box estimation by regression forest}

We obtain patches from an unknown input CT volume.
The patches are input to the root of each tree.
In each split node, the feature values are calculated and thresholded using the recorded values.
The result is used to select a child node to go through.
When the patch reaches a leaf node, the average of the normal distribution, which is recorded in the leaf node, is given as the output of the regression tree.
The output is offset from the patch position.
The sum of the offset and the patch position is the estimated position of the face of a bounding box.
The average of the estimated positions of all the patches is the final estimation result of the face position of a bounding box.

\subsection{Directed structure specific (DSS) volume}

\begin{figure}[tb]
\begin{center}
\begin{tabular}{ccc}
\includegraphics[width=0.3\textwidth]{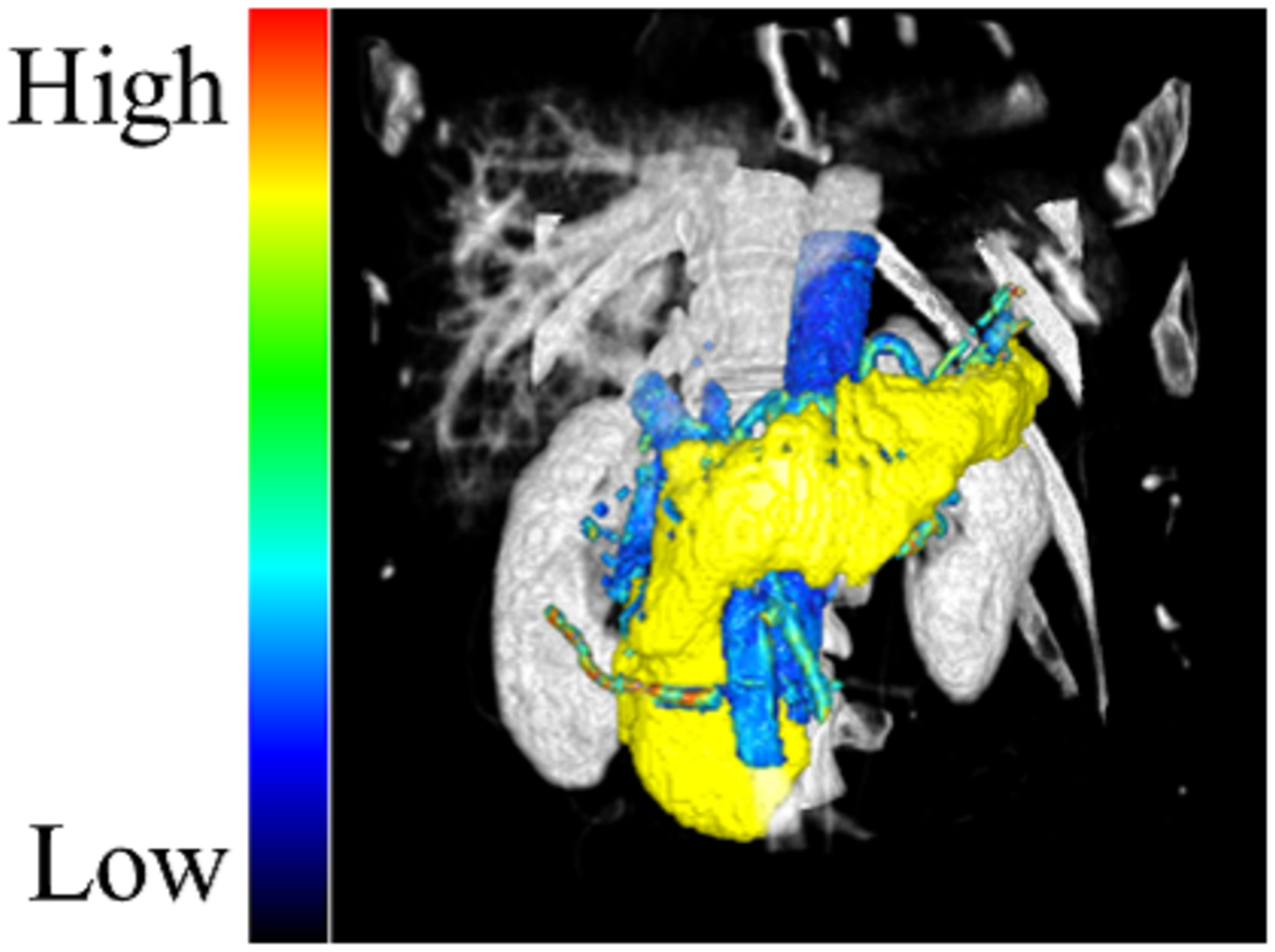}
&
\includegraphics[width=0.29\textwidth]{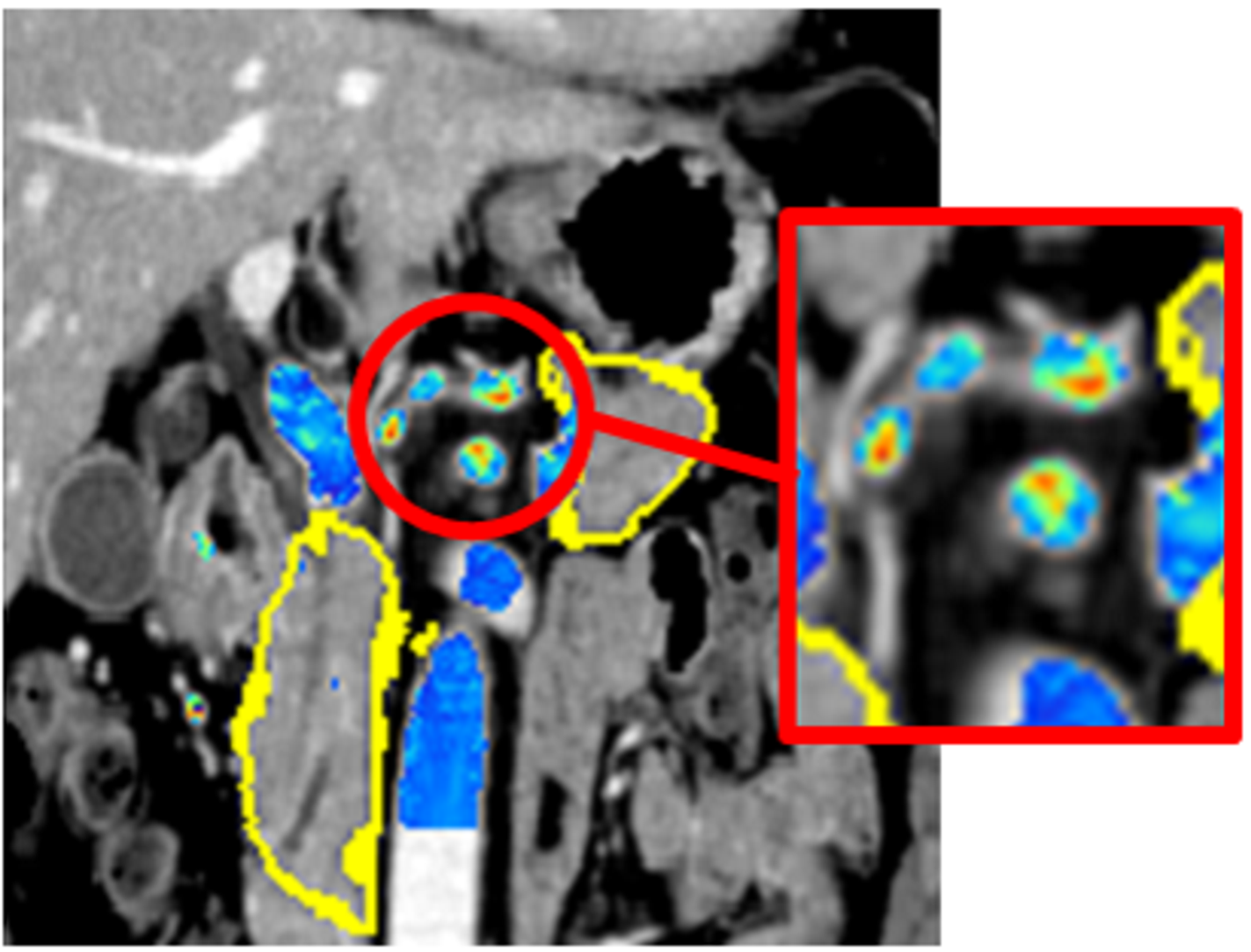}
&
\includegraphics[width=0.28\textwidth]{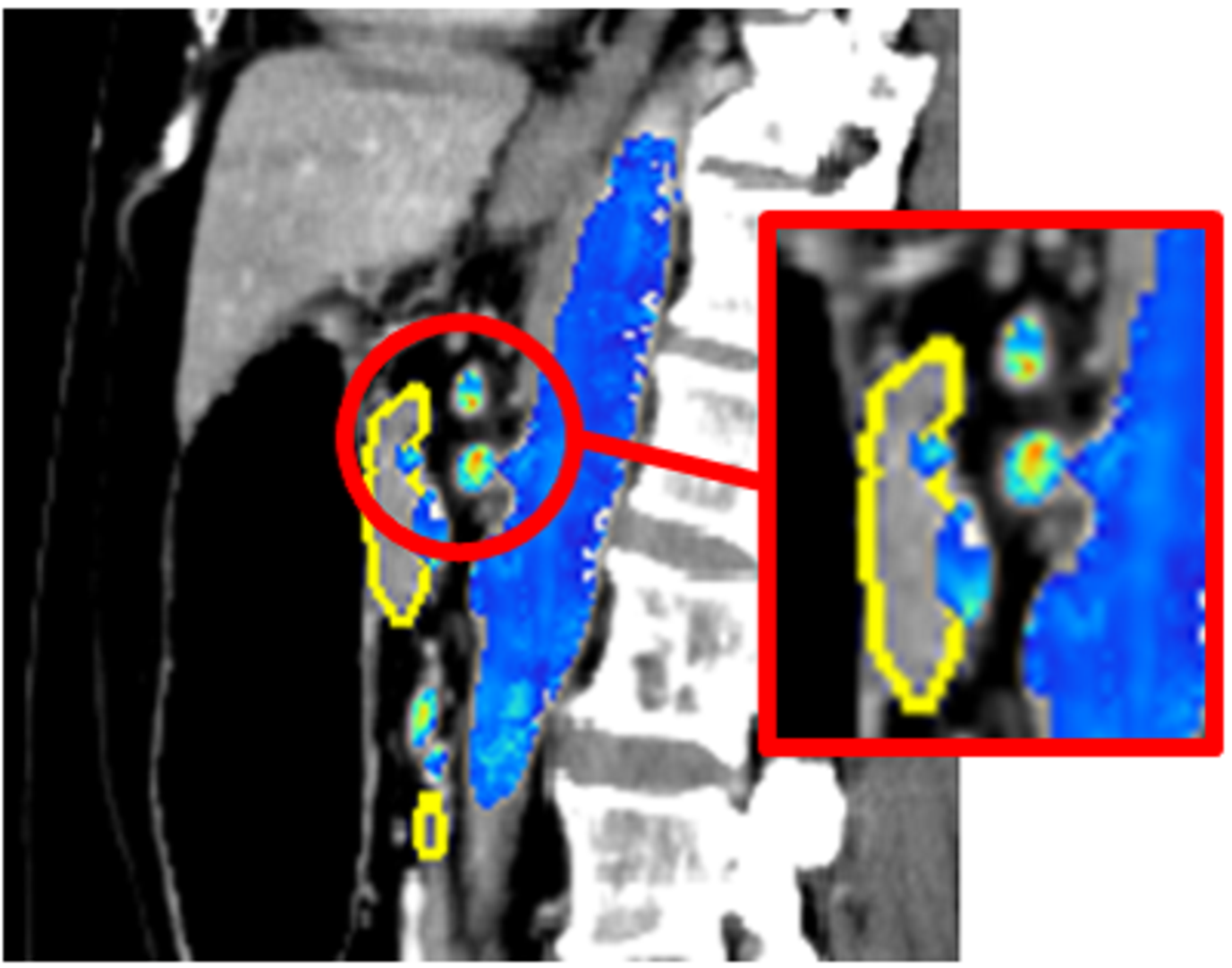}\\
(a) & (b) & (c)
\end{tabular}
\caption{An example of the DSS volume overlaid on CT volume. Pancreas is shown in yellow. DSS volume intensities are indicated by colors: red is high, and blue is low. (a) 3D view, (b) coronal, and (c) sagittal slice views. DSS volume has high values in blood vessels, especially in SV, which is indicated by red circles.}
\label{fig:dssvolume}
\end{center}
\end{figure}

A probabilistic atlas of the pancreas is generated from volumes in the training database that have high similarities to the input CT volume.
We generate DSS volumes for the atlas selection.
Blood vessels are circulating around the pancreas.
The splenic vein (SV), whose position has high correlation with the pancreas position.
We generate a DSS volume that stores high values in the SV.
An example of DSS volume is shown in Fig. \ref{fig:dssvolume}.

The multi-scale line structure enhancement filter\cite{Sato98} is applied to a CT volume.
We generate DSS volume as follows: 
\begin{eqnarray}
M({\bf x}) = \left\{ \begin{array}{ll}
w \cdot \max_{1 \leq k \leq m} \{ \sigma^{2}_{k} \cdot \lambda_{123}({\bf x};\sigma_{k}) \}, \ \ \ & {\rm if} \ | {\bf e}^{\rm T}_{1}{\bf u}_{z} | \leq \tau, \\
\max_{1 \leq k \leq m} \{ \sigma^{2}_{k} \cdot \lambda_{123}({\bf x};\sigma_{k}) \}, \ \ \ & {\rm otherwise},
\end{array} \right.
\label{eq:dssv}
\end{eqnarray}
where $\sigma_{k} = k\sigma_{1}$, $\sigma_{1}$ is the size of the enhancement target line and $\lambda_{123}$ is the response of the line structure enhancement filter.
${\bf e}_{1}$ is an eigenvector that corresponds to the largest eigenvalue of the Hessian matrix calculated at ${\bf x}$, ${\bf u}_{z}$ is the head to foot direction unit vector, $\tau$ is a threshold of the vector directions, and $m$ is the scale step number.
At the voxels in the line structures, ${\bf e}_{1}$ becomes the direction along the line, meaning that ${\bf e}_{1}$ shows the blood vessel directions.
The SV direction is almost perpendicular to the head to foot direction.
We give higher weight in the calculation of Eq. \ref{eq:dssv} at the voxels in the SV using the condition of the ${\bf e}_{1}$ and ${\bf u}_{z}$ directions.

\subsection{Patient-specific probabilistic atlas generation}
\label{ssec:atlasgeneration}

\begin{figure}[tb]
\begin{center}
\includegraphics[width=0.95\textwidth]{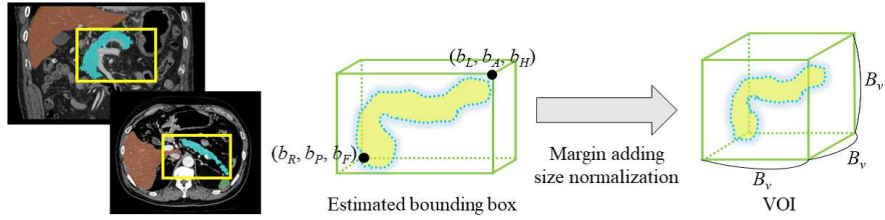}
\caption{Pancreas bounding box is estimated on CT volume. Fixed margin is added to it and its size is normalized to create VOI.}
\label{fig:voi}
\end{center}
\end{figure}

We generate normalized pancreas bounding boxes of the training database and the input CT volume.
The bounding box of each data in the database is estimated using a method described in Section \ref{ssec:localization}.
We add a fixed size of the margin, which is selected so that bounding box covers most of the pancreas region, to the bounding box for reducing the effect of estimation error.
We crop the CT volume and a corresponding manually segmented pancreas in the bounding box with the margin from the data in the database. %input ctのことも分かるように書く
The sizes of the cropped CT volume and the manually segmented pancreas are normalized to an image size of $B_{v}^{3}$ voxels and a voxel size of $B_{s}^{3}$ mm.
We call a set of a cropped CT volume and the corresponding cropped manually segmented pancreas in the database a {\it database VOI}.
Also, the input CT volume is cropped and we refer to this as {\it input VOI}.
A schematic illustration of VOI generation is shown in Fig. \ref{fig:voi}.

Next we explain how we select the database VOIs for atlas generation.
DSS volumes are generated from the CT volumes in the database VOIs and the input VOI.
All of the CT volumes in the database VOIs are registered to the CT volume in the input VOI using the MRF-based non-rigid registration method\cite{Glocker08}.
The corresponding manually segmented pancreases and DSS volumes are also deformed by the registration process.
%We calculate the similarities between the DSS volumes of the database VOIs and the input VOI by a similarity measure known as the zero-mean normalized cross-correlation (ZNCC)\cite{Brown92}.
We calculate the similarities between the DSS volumes of the database VOIs and the input VOI by the zero-mean normalized cross-correlation (ZNCC)\cite{Brown92}.
We select $N_{s}$ database VOIs that have high similarities with input VOI.

A patient-specific probabilistic atlas for the input CT volume is generated from the selected database VOIs.
The atlas is obtained by
\begin{equation}
M_{{\bf p} \in V^{(l)}} = \frac{ \sum^{N_{s}-1}_{i=0} z_{i} \cdot \delta(L^{i}_{\bf p},l) }{ \sum^{N_{s}-1}_{i=0} z_{i} },
\end{equation}
where $M_{{\bf p} \in V^{(l)}}$ is a probabilistic atlas of organ label $l$ for voxel ${\bf p}$ in input VOI $V$.
$z_{i}$ is the weight of the $i$-th database VOI.
Here we use the ZNCC value as the weight.
$\delta(L^{i}_{\bf p},l)$, which is a delta function, is described as
\begin{eqnarray}
\delta(l,l') = \left\{ \begin{array}{ll}
1, & \ \ \  {\rm if} \ l=l', \\
0, & \ \ \ {\rm otherwise},\
\end{array} \right.
\end{eqnarray}
where $L^{i}_{\bf p}$ is an organ label of voxel ${\bf p}$ in the $i$-th database VOI.

\subsection{Pancreas segmentation}

For the input CT volume, a rough segmentation is first performed using the patient-specific probabilistic atlas generated in Section \ref{ssec:atlasgeneration}.
We used maximum a posteriori estimation in the rough segmentation\cite{Karasawa15}.
Precise segmentation is performed using the graph-cut method\cite{Boykov01} to refine rough segmentation result.

\section{Experiments and Discussion}

\begin{figure}[tb]
\begin{center}
\begin{tabular}{ccc}
\includegraphics[width=0.35\textwidth]{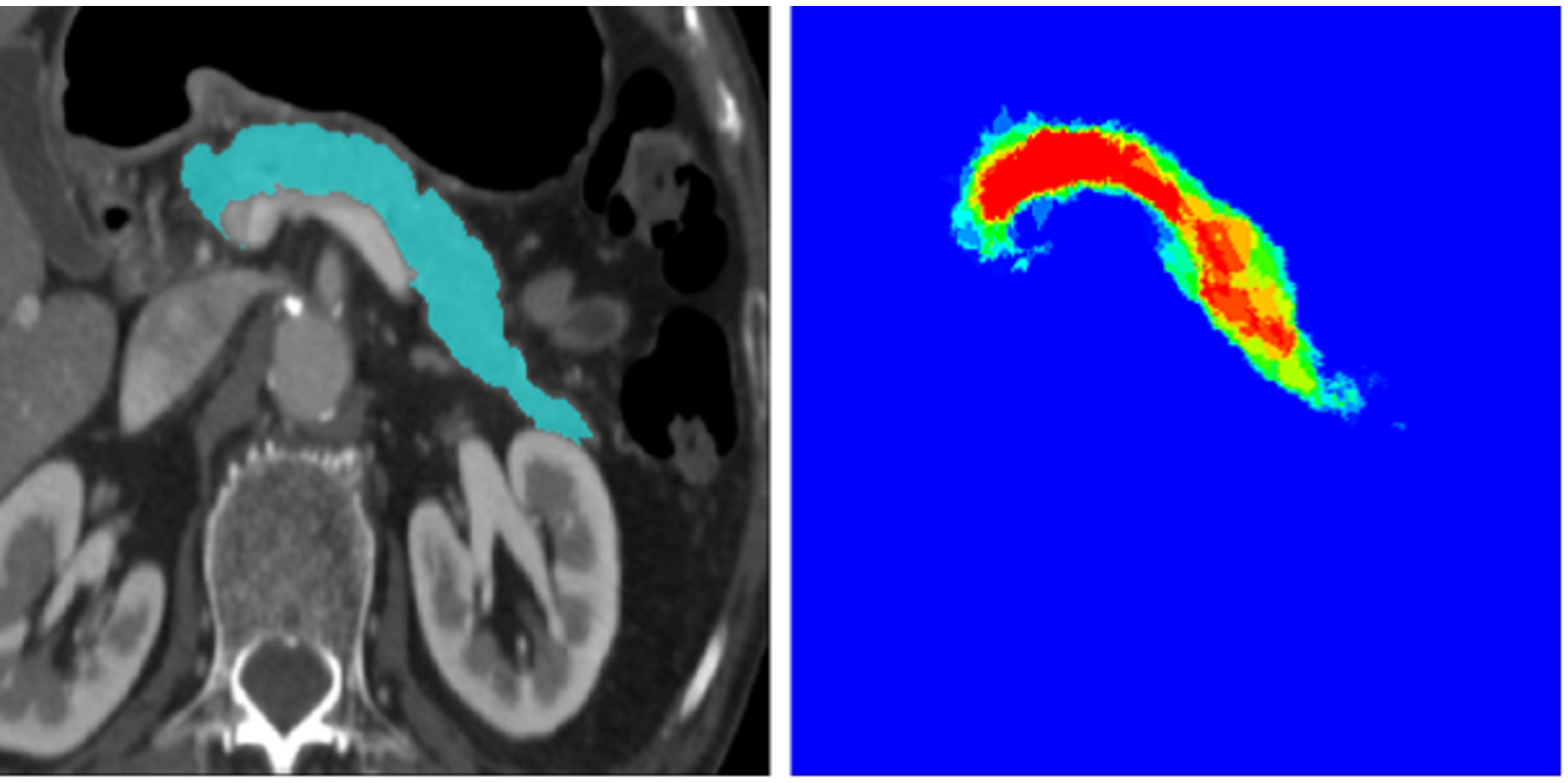}
& \ \ \ &
\includegraphics[width=0.35\textwidth]{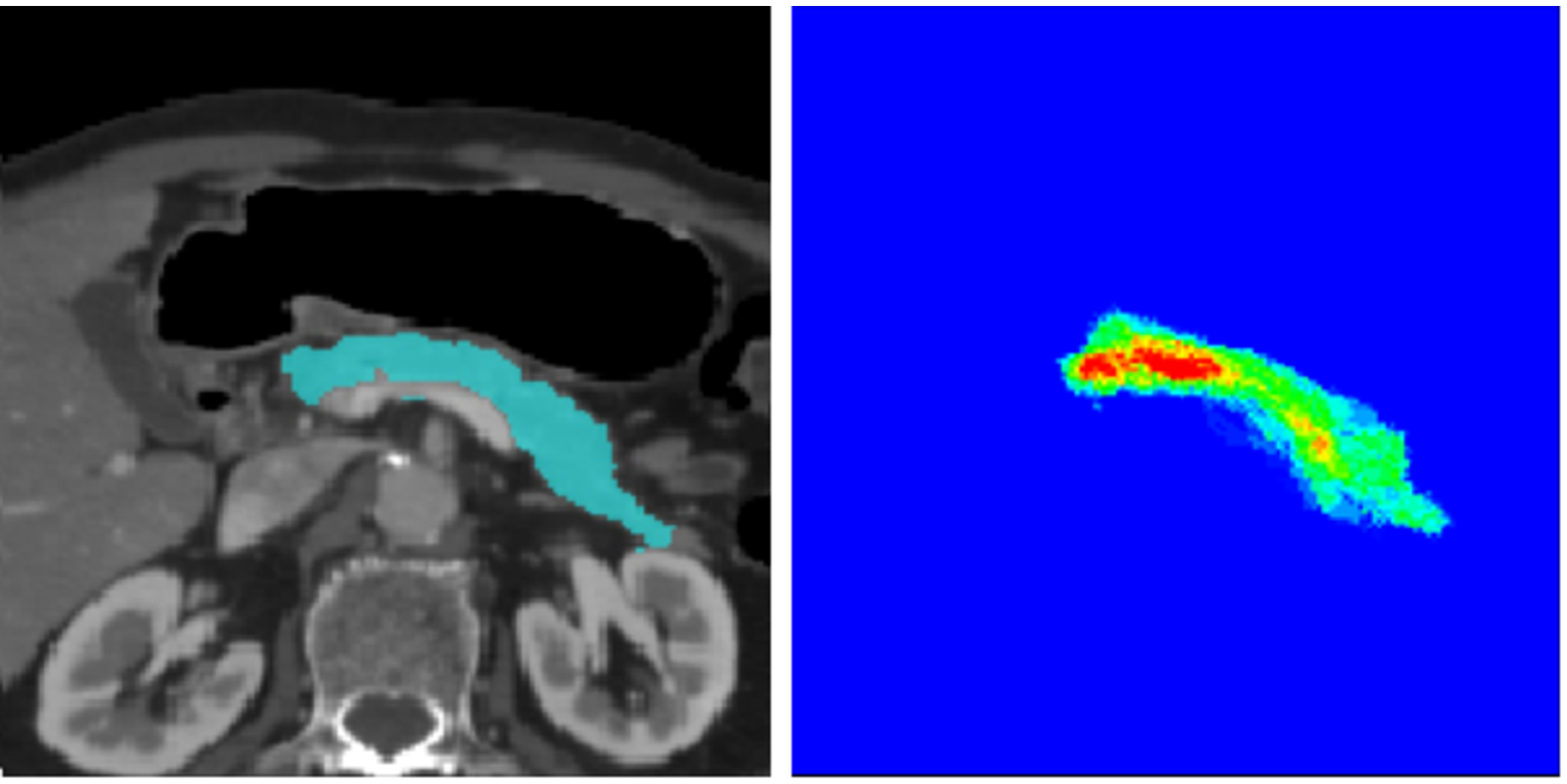}\\
(a) Proposed (JI 86.4\%, DICE 92.7\%) & & (b) Previous (JI 68.4\%, DICE 81.3\%)\\
\includegraphics[width=0.35\textwidth]{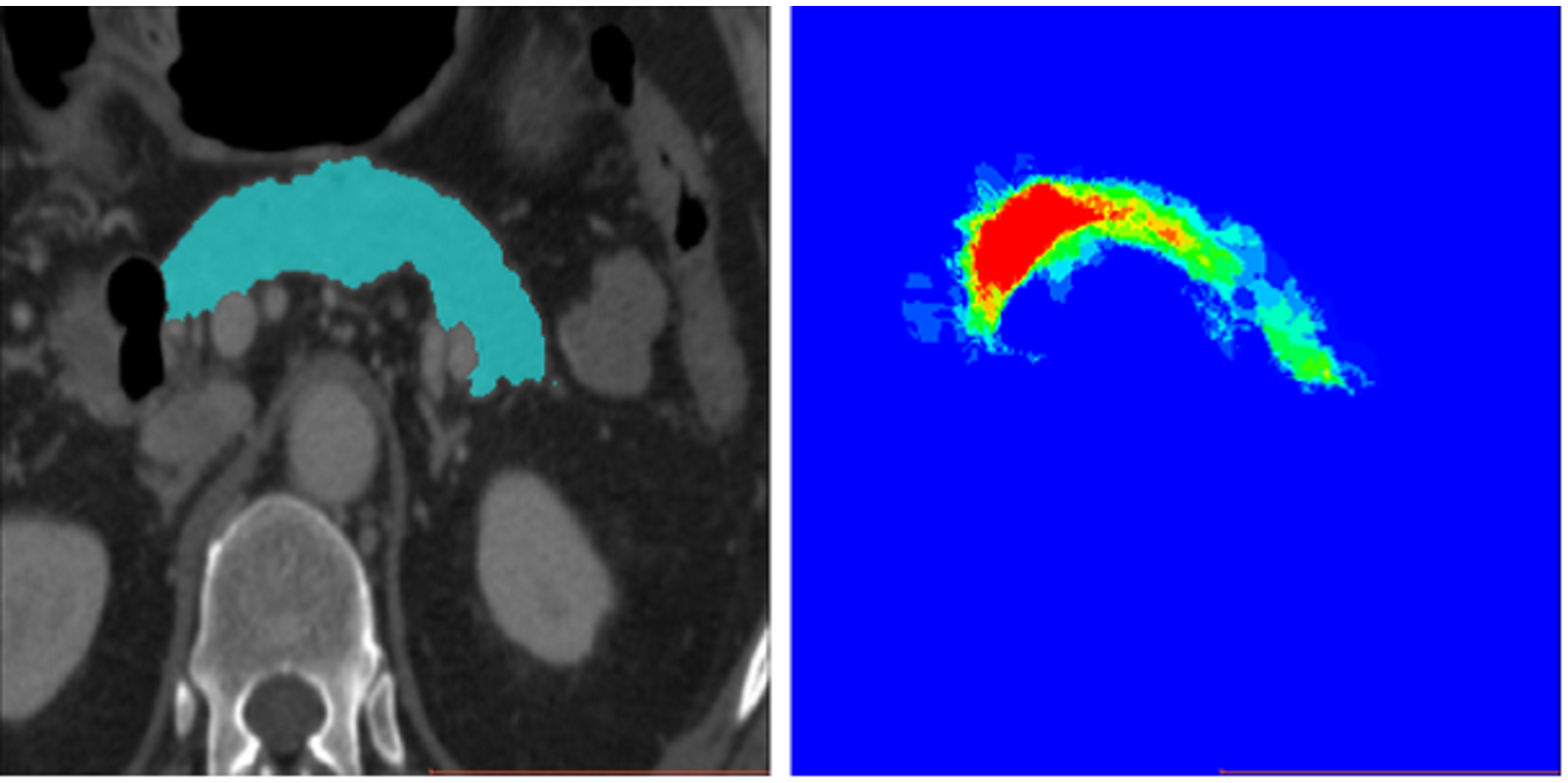}
& \ \ \ &
\includegraphics[width=0.35\textwidth]{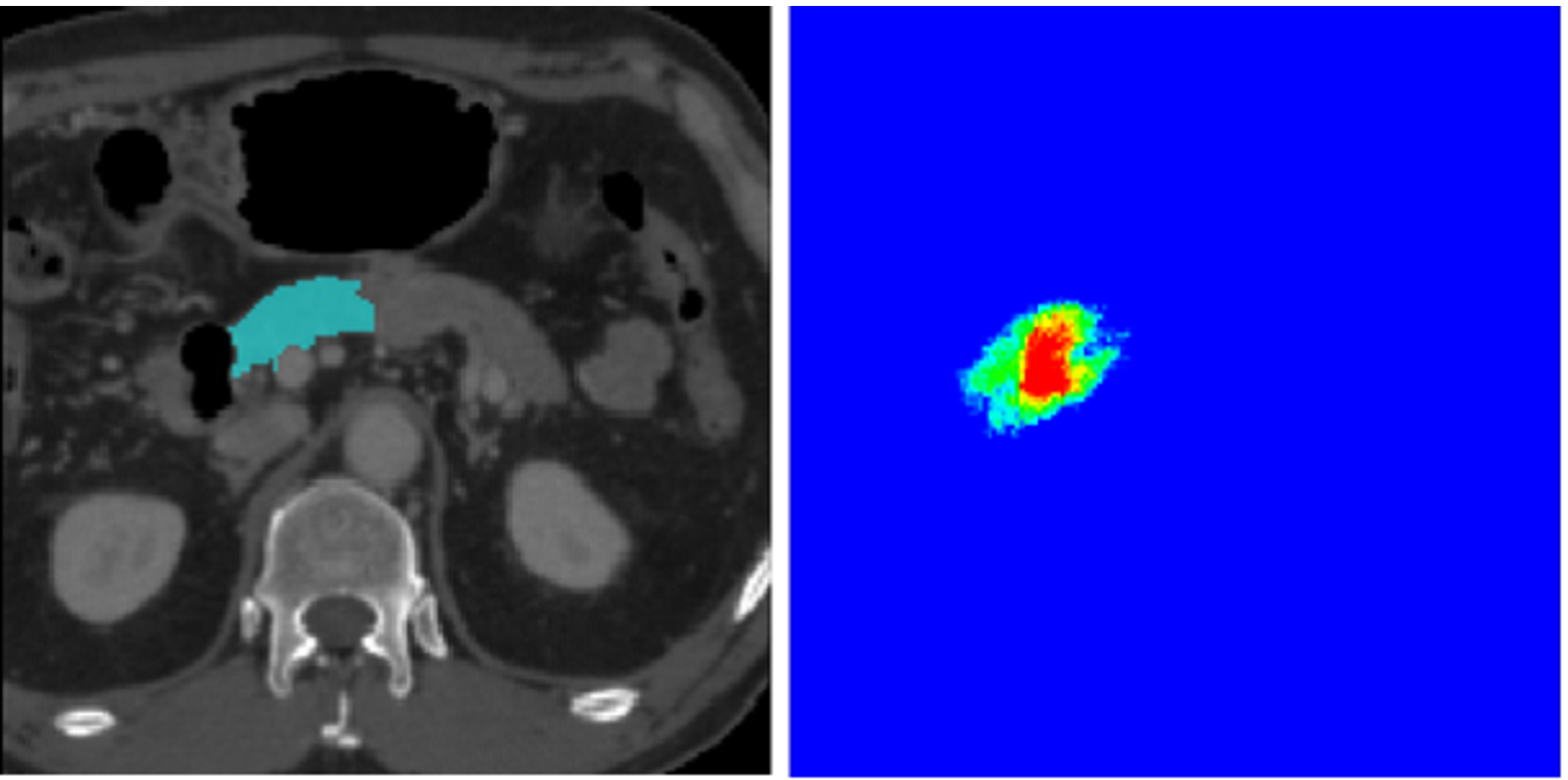}\\
(c) Proposed (JI 85.1\%, DICE 91.9\%) & & (d) Previous (JI 48.4\%, DICE 65.2\%)\\
\includegraphics[width=0.35\textwidth]{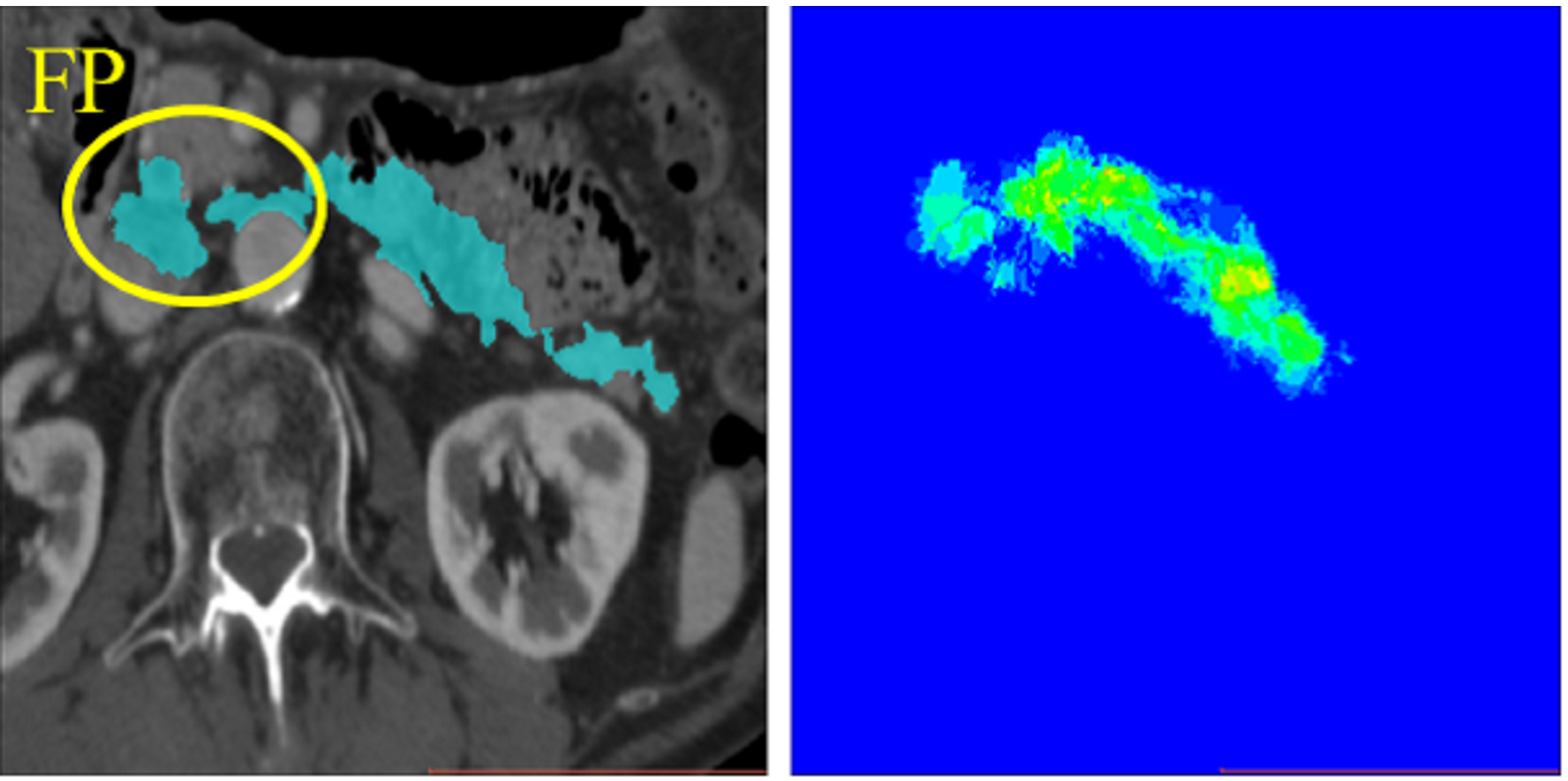}
& \ \ \ &
\includegraphics[width=0.35\textwidth]{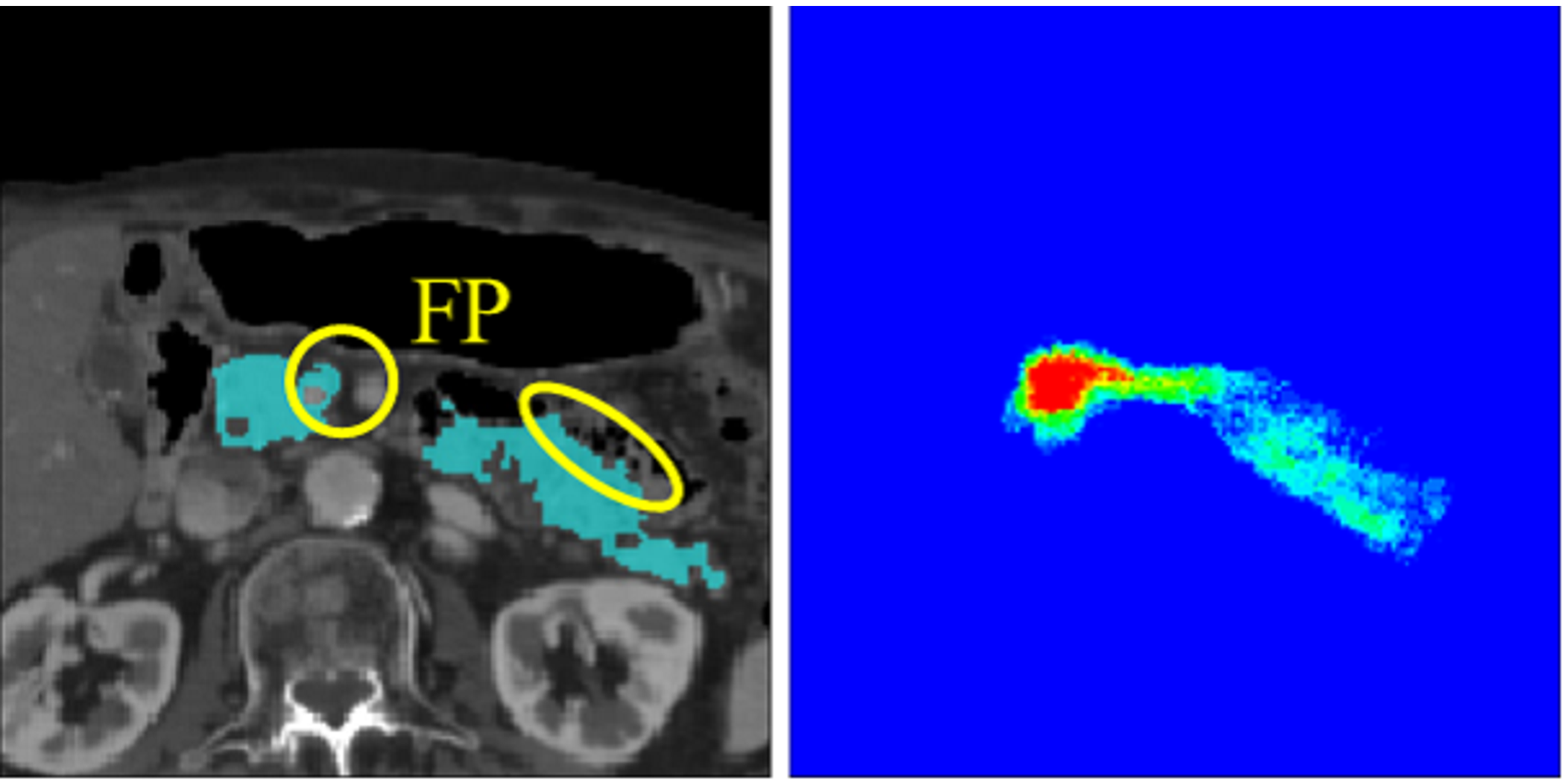}\\
(e) Proposed (JI 10.4\%, DICE 18.9\%) & & (f) Previous (JI 43.0\%, DICE 60.2\%)
\end{tabular}
\caption{Examples of pancreas segmentation (left) and corresponding probabilistic atlas (right) of proposed and previous\cite{Karasawa15} methods. These are slice images of calculated VOIs. Note that their aspect ratios are different from original CT images because size normalization was applied. Segmented pancreases are colored blue and overlaid on axial CT slices. In probabilistic atlases, red means high and blue means low values. Three pairs (a)--(b), (c)--(d), and (e)--(f) show same cases and same axial slices.}
\label{fig:result}
\end{center}
\end{figure}

We evaluated the proposed method using 147 cases of abdominal CT volumes.
The following are the acquisition parameters of the CT volumes: image size: $512 \times 512$ voxels, slice number: 263--1061 slices, pixel spacing: 0.546--0.820 mm, and slice spacing: 0.40--0.80 mm.
The ground truth pancreas, liver, spleen, and kidneys regions were semi-automatically made by three trained researchers.
The ground truth region of each case was made by one of three researchers.
All ground truth regions were checked by a medical doctor.
The parameters of the method were experimentally selected as $p=25, n_{F}=40, n_{H}=500, n_{min}=20, D=15, B_{v}=256, B_{s}=1.0, N_{s}=20, w=2.0, m=7, \tau=0.25,$ and $\sigma_{1}=1.0$.
We evaluated the proposed method by leave-one-out cross validation by the following evaluation metrics: Jaccard Index (JI) and Dice Overlap (DICE).
Table \ref{tab:result} compares the accuracies of the proposed and previous methods.
Figure \ref{fig:result} shows the segmentation results.
The average computation times of the pancreas localization and segmentation were 44.7 seconds and about three hours per case.

\begin{table}[tb]
\begin{center}
\caption{Accuracies of proposed and previous pancreas segmentation methods.}
\label{tab:result}
\begin{tabular}{|c|c|c|c|} 
\hline
Method & Data number & JI (\%) & DICE (\%) \\ \hline \hline
Proposed & 147 & 62.1$\pm$16.6 & 75.1$\pm$15.4 \\ \hline
Okada et al.\cite{Okada08} & 86 & 59.2 & 71.8 \\ \hline
Chu et al.\cite{Chu13} & 100 & 54.6$\pm$15.9 & 69.1$\pm$15.3 \\ \hline
Wolz et al.\cite{Wolz13} & 150 & 55.0$\pm$17.1 & 69.6$\pm$16.7 \\ \hline
Karasawa et al.\cite{Karasawa15} & 150 & 61.6$\pm$16.6 & 74.7$\pm$15.1 \\ \hline
Tong et al.\cite{Tong15} & 150 & 56.9$\pm$15.2 & 71.1$\pm$14.7 \\ \hline
Saito et al.\cite{Saito16} & 140 & 62.3$\pm$19.5 & 74.4$\pm$20.2 \\ \hline
\end{tabular}
\end{center}
\end{table}

%\section{Discussion and Conclusions}

The most important contribution of our method is full automation of the pancreas segmentation process while achieving comparable accuracy to the previous state-of-the-art method\cite{Karasawa15}.
The results show that the segmentation accuracy of the proposed method exceeded most of the previous methods.
Our fully automated method is quite useful for various medical assistance applications.
The proposed method tackles the spatial and shape variations of the pancreas via regression forest-based localization and patient-specific atlases.
Although we automated all of the segmentation processes, segmentation results were superior to the other state-of-the-art methods in the DICE.

Previous methods\cite{Chu13,Wolz13,Karasawa15} utilized manually segmented regions or the organ positions (other than the pancreas) for pancreas segmentations.
They use this information for pancreas localization.
Our method does not require any manual input or additional information for atlas localization; this is its great advantage.
In Figs. \ref{fig:result} (a) and (c), the calculated VOIs included the pancreas regions in their centers.
These demonstrate that the pancreas localization successfully works.
In these cases, much fat tissues exist among organs.
Fat tissues have lower CT values compared to organs.
Existence of fat tissue clarifies the organ contours.
We used the intensity difference-based feature value (Eq. \ref{eq:featurevalue}) in the pancreas localization.
Clear organ contours make accurate localization of the pancreas.
Compared to them, the pancreas touches the surrounding organs in Fig. \ref{fig:result} (e).
This makes the organ contours obscure and reduces the segmentation accuracy.

Results of the previous method\cite{Karasawa15} are shown in Figs. \ref{fig:result} (b), (d), and (f).
The previous method generates pancreas VOIs using the manually specified liver positions.
VOIs of the previous method correctly included the pancreas in most cases.
However, probabilistic atlases of the previous method were not good.
Atlas that have high and low likelihoods inside and outside the pancreas contributes accurate segmentations.
The previous method used line structure enhanced volumes for probabilistic atlas generation.
We newly introduced DSS volume for probabilistic atlas generation.
In the DSS volume, the SV and other blood vessels were enhanced.
The enhancement result was quite effective for selecting similar CT volumes from the database in the atlas generation.
In Figs. \ref{fig:result} (a) and (c), generated atlases have high values in pancreas regions.
The good atlases contributed segmentation accuracy improvements.

This paper proposed a fully automated pancreas segmentation method from a CT volume using a regression forest technique and patient-specific atlas generation.
In the atlas generation, we introduced a new similarity based on blood vessel position and direction.
Experiments using 147 CT volumes outperformed the other state-of-the-art methods.
Future work includes the utilization of new feature values that can capture the local appearances of organs in pancreas localization and remove the misalignments of bounding box faces.
Assessment of inter- and intra-observer variability of the ground truth data is also required.

\subsubsection*{Acknowledgments.} 
%The authors thank *****, *****, *****, *****, *****, *****, and *****.
Parts of this research were supported by the MEXT, the JSPS KAKENHI Grant Numbers 25242047, 26108006, 26560255, and the JSPS Bilateral International Collaboration Grants.

\end{document}